%% file: spitzer_error_metrics.tex
\documentclass[conference]{IEEEtran}
\usepackage[comma, numbers]{natbib}
\usepackage{amsmath,amssymb,amsfonts}
\usepackage{algorithmic}
\usepackage{graphicx}
\usepackage{textcomp}
\usepackage{xcolor}
\def\BibTeX{{\rm B\kern-.05em{\sc i\kern-.025em b}\kern-.08em
    T\kern-.1667em\lower.7ex\hbox{E}\kern-.125emX}}

\usepackage{hyperref}

\include{commands}

\begin{document}

\title{Rotational Error Metrics for Quadrotor Control}

\author{\IEEEauthorblockN{Alexander Spitzer}
\IEEEauthorblockA{\textit{Robotics Institute} \\
\textit{Carnegie Mellon University}\\
Pittsburgh, PA, USA \\
spitzer@cmu.edu}
\and
\IEEEauthorblockN{Nathan Michael}
\IEEEauthorblockA{\textit{Robotics Institute} \\
\textit{Carnegie Mellon University}\\
Pittsburgh, PA, USA \\
nmichael@cmu.edu}
}

\maketitle

\input{abstract}


\bstctlcite{IEEEexample:BSTcontrol}

\section{Introduction}

We analyze and compare various attitude error metrics that have been proposed for use in quadrotor attitude controllers.

\begin{figure}
  \begin{center}
  \includegraphics[width=9cm]{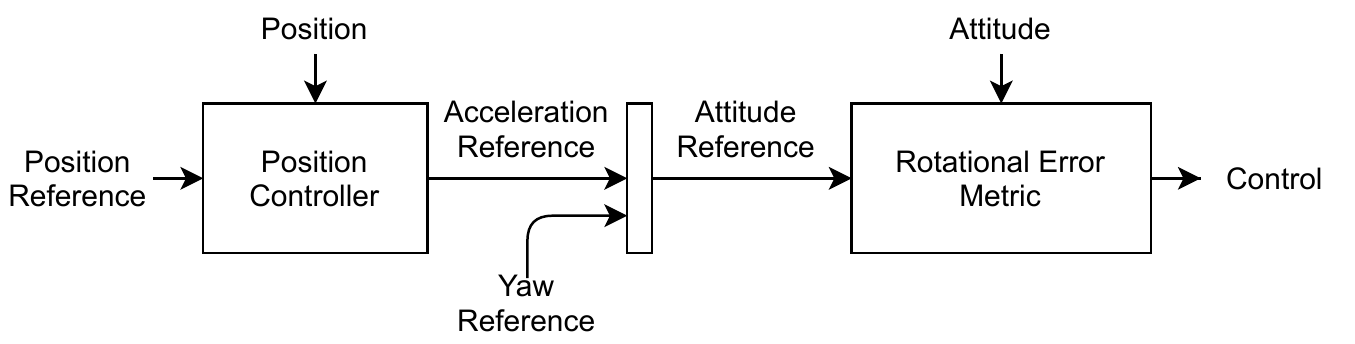}
  \caption{The standard cascaded quadrotor controller architecture. The position controller computes a desired acceleration, which is then combined with the yaw reference to generate an attitude reference for the attitude controller. The attitude controller then uses a rotational error metric to compute the controls.}
  \label{fig:control_diag}
  \end{center}
\end{figure}

The most common control structure for multirotors is cascaded, shown in Figure~\ref{fig:control_diag}, where a position controller generates a desired attitude $R_{\text{des}}$ that an attitude controller then attempts to track.

For brevity, we consider regulation, not trajectory tracking, but all of the results here can be extended to work for trajectory tracking, where the desired angular velocity $\omega_{\text{des}}$ and the desired angular acceleration $\alpha_{\text{des}}$ are nonzero.

The structure of quadrotor attitude controllers that we consider is a PD controller on rotational error, as defined by a rotational error metric $e_R$, scaled by a diagonal attitude gain matrix $K_R$, and angular velocity $\omega$, scaled by a diagonal angular velocity gain matrix $K_\omega$.

\newcommand{\Rdes}{R_{\text{des}}}
\newcommand{\zdes}{z_{\text{des}}}

\begin{align}
  \alpha = -K_R e_R(R, \Rdes) - K_\omega \omega
\end{align}

Several works in the literature have noted that, since the quadrotor's position dynamics depend solely on its tilt, or body $z$-axis, it makes sense to prioritize vehicle tilt over vehicle yaw [\citenum{brescianini_nonlinear_2013,faessler_automatic_2015,mueller_multicopter_2018,kooijman_trajectory_2019,gamagedara_geometric_2019}].
In this work, we confirm this, and show that such metrics amount to inducing a response in the direction of the cross product between the current and desired body $z$-axes, along with a response around the body $z$-axis.
We show that this decomposition linearizes the vehicle error response better than the traditional attitude controller, which uses the full rotation error.
Following straight paths in the presence of large tracking error can be beneficial for overall control performance.


\section{Rotational Error Metrics}

\subsection{Preliminaries}

In this section, we place various controllers used in the literature in the framework described above.
First, as in \citet{mueller_multicopter_2018}, we define rotational error $R_e$ as

\begin{align}
  R_e = \Rdes^\top R
\end{align}

and its corresponding angle and axis as $\rho_e$ and $n_e$.

The vehicle attitude $R$ transforms vectors from the body frame $\mathcal B$ into the fixed frame, and is represented using a matrix whose columns $x$, $y$, and $z$ are the coordinate frames axes of the body.

\begin{align}
  R = \begin{pmatrix} x & y & z \end{pmatrix}
\end{align}

We define the \textit{thrust vector} error $R_{tv}$ as the rotation between the $z$-axes of the current and desired attitude, along with its associated angle $\rho_{tv}$ and axis $n_{tv}$.
This rotation can be defined via its angle and axis in terms of the full rotation error $R_e$ as shown below \citep{mueller_multicopter_2018}.

\begin{align}
  \rho_{tv} &= \cos^{-1}(e_3^\top R_e e_3) \\
  n_{tv} &= \frac{(R_e^\top e_3) \times e_3 }{|| (R_e^\top e_3) \times e_3 || }
\end{align}

To help visualize $n_{tv}$, note that it is just the normalized cross product of the desired $z$-axis $\zdes^{\mathcal{B}} = R^\top\Rdes e_3$ with the current $z$-axis of the body $z^{\mathcal{B}} = e_3$, expressed in the body frame.
As a result, $n_{tv}^\top e_3 = 0$.

\begin{align}
  n_{tv} = \frac{
                  \zdes^{\mathcal{B}} \times z^{\mathcal{B}}
                }
                { ||
                  \zdes^{\mathcal{B}} \times z^{\mathcal{B}}
                || }
\end{align}

We define the yaw error $R_{\text{yaw}}$ as the remaining error around the body $z$-axis, i.e. the rotation needed to align the current rotation to the desired rotation after aligning the $z$ axes.

\begin{align}
  R_{\text{yaw}} = R_e R_{tv}^{-1}
\end{align}

$\rho_{\text{yaw}}$ represents the angle associated with $R_{\text{yaw}}$ and $n_{\text{yaw}}$ represents the axis, which is either $e_3 = (0, 0, 1)^\top$ or $-e_3$, and thus perpendicular to $n_{tv}$.
\subsection{Full Rotation Metrics}

A rotational error metric directly in $SO(3)$ is used in [\citenum{lee_geometric_2010,mellinger_minimum_2011,goodarzi_geometric_2013}].
As shown in \cite{mueller_multicopter_2018}, this metric is proportional to the sine of the angle error $\rho_e$ and in the direction of the shortest path $n_e$.

\begin{align}
  \label{eq:skew}
  e_R^{\textbf{A}}(R, \Rdes) = \frac12\left(R_e - R_e^\top\right)^\vee = \sin\rho_e n_e
\end{align}

\citet{lee_exponential_2012} notes the issues with using a rotational error metric that is proportional to the sine of the error, namely that the response is strongest at an error of $90^\circ$ and weakens as the error increases, reaching zero at an error of $180^\circ$.
While this ensures smoothness of the response, it results in slow convergence when the error is high.
\citet{lee_exponential_2012} proposes rescaling the rotational error. \footnote{We add a factor of 2 to match the linearization around $\rho_e = 0$ of \eqref{eq:skew}.}

\begin{align}
  \label{eq:scaled_skew}
  e_R^{\textbf{B}}(R, \Rdes) = \frac{2}{\sqrt{1 + \tr\left(R_e\right)}}\sin\rho_e n_e
\end{align}

Using the fact that $\tr\left(R_e\right) = 2\cos\rho_e + 1$ \citep{mueller_multicopter_2018} and the sine half angle identity $\sin\left(\frac{\theta}{2}\right) = \sqrt{\frac{1 - \cos\theta}{2}}$,
\eqref{eq:scaled_skew} can be rewritten as

\begin{align}
  \label{eq:sinhalf_full}
  e_R^{\textbf{B}}(R, \Rdes) = 2\sin\left(\frac{\rho_e}{2}\right) n_e.
\end{align}

\eqref{eq:sinhalf_full} is mathematically equivalent to the rotational error used in \citet{fresk_full_2013}, which is the axis component of the error quaternion. The axis component of a quaternion is the sine of the half angle multiplied by the axis.

Similarly, one can use an error response that is directly proportional to the angle.

\begin{align}
  \label{eq:angle_full}
  e_R^{\textbf{C}}(R, \Rdes) = \rho_e n_e.
\end{align}

This is likely not as widely used, since unlike \eqref{eq:skew} and \eqref{eq:scaled_skew}, evaluating \eqref{eq:angle_full} from a rotation matrix or quaternion attitude representation requires inverse trigonometric function evaluations, which may be expensive on an embedded platform.

\subsection{Thrust Vector -- Yaw Decomposition Metrics}

In this section we will characterize the attitude controls used by works that decompose attitude into a thrust vector and yaw, also known as ``reduced attitude control'', with an emphasis on the control around the body $x$ and $y$ axes.
Define, $\omega_{XY} = z \times \dot z = \omega - \left(\omega^\top z\right) z$.

\citet{kooijman_trajectory_2019} decomposes the attitude representation into $\mathcal S_2 \times \mathcal S_1$, which amounts to controlling the thrust vector direction, an element of $\mathcal S_2$, independently from the rotation around the thrust vector, an element of $\mathcal S_1$.
The control response around the $x$ and $y$ axes\footnote{For simplicity, we omit the yaw controller from \citet{kooijman_trajectory_2019}.} in \citet{kooijman_trajectory_2019} is given as

\begin{align}
  \label{eq:s2s1_omxy}
  \omega_{XY}^{\mathcal B} &= R^\top ( z \times u_v ) \\
              &= e_3 \times (R^\top u_v) \\
              &= \begin{pmatrix} -y^\top u_v \quad x^\top u_v \quad 0 \end{pmatrix}^\top,
\end{align}

with $u_v$ defined as

\begin{align}
  \label{eq:uvdef}
  u_v =
    \begin{cases}
    k_1 \zdes &  \rho_{tv} \leq \frac\pi2 \\
    \frac{k_1}{\sqrt{1 - (z^\top\zdes)^2}} \zdes =
    \frac{k_1}{\sin\rho_{tv}} \zdes &  \frac\pi2 < \rho_{tv} < \pi.
  \end{cases}
\end{align}

Substituting \eqref{eq:uvdef} into \eqref{eq:s2s1_omxy}, we get

\begin{align}
  \label{eq:s2s1_omxy2}
  \omega_{XY}^{\mathcal B} =
  \begin{cases}
    -k_1 \zdes^{\mathcal B} \times z^{\mathcal B} = -k_1 \sin\rho_{tv} n_{tv} &  \rho_{tv} \leq \frac\pi2 \\
    -k_1 \frac{\zdes^{\mathcal B} \times z^{\mathcal B}}{\sin\rho_{tv}} = -k_1 n_{tv} & \frac\pi2 < \rho_{tv} < \pi.
  \end{cases}
\end{align}

Assuming a P controller for angular velocity, from \eqref{eq:s2s1_omxy2} we can extract the rotational error metric $e_R$ as

\begin{align}
  \label{eq:s2s1_re}
  e_R^{\textbf{D}}(R, \Rdes) =
   \begin{cases}
     \sin\rho_{tv} n_{tv} & \rho_{tv} \leq \frac\pi2 \\
     n_{tv} & \frac\pi2 < \rho_{tv} < \pi.
   \end{cases}
\end{align}

\eqref{eq:s2s1_re} has the interesting property that the magnitude of the response is maximum at an angle of $\rho_{tv} = \frac\pi2$ and remains constant as the angle increases from $\frac\pi2$ to $\pi$.
For $\rho_{tv} \leq \frac\pi2$, \eqref{eq:s2s1_re} is equivalent to \eqref{eq:skew} for axis-aligned errors.

\citet{brescianini_tilt-prioritized_2020} also decomposes the thrust vector and the yaw, but does so using two quaternions.
The final control law is a sum of a quaternion error representing the error in the thrust direction and a quaternion error representing the error in the angle around the body $z$-axis.
The use of quaternions implies that the rotational error metric is proportional to the sine of the half angle for both the thrust vector error and the yaw error.
The resulting rotational error metric is shown below, with again, a factor of 2 added to match linearizations with the other metrics.

\begin{align}
  \label{eq:quat_decomp}
  e_R^{\textbf{E}}(R, \Rdes) = 2\sin\left(\frac{\rho_{tv}}{2}\right) n_{tv} + 2 \sin\left(\frac{\rho_{\text{yaw}}}{2}\right) n_{\text{yaw}}
\end{align}

\eqref{eq:quat_decomp} is equivalent to \eqref{eq:sinhalf_full} for axis-aligned errors.

\citet{mueller_multicopter_2018} proposes a new rotational error metric that is effectively a linear interpolation between \eqref{eq:angle_full} and only controlling the thrust vector. The rotational error metric is shown below, with $\alpha_{\text{yaw}} \in [0, 1]$ used to weight the control of the yaw angle.

\begin{align}
  \label{eq:mueller4}
  e_R^{\textbf{F}}(R, \Rdes) = \alpha_{\text{yaw}} \rho_e n_e + (1 - \alpha_{\text{yaw}}) \rho_{tv} n_{tv}
\end{align}

Unlike, \eqref{eq:s2s1_re} and \eqref{eq:quat_decomp}, \eqref{eq:mueller4} doesn't decouple the yaw control from the thrust vector control for $\alpha_{\text{yaw}} > 0$.
As we will show below, this leads to suboptimal performance in certain situations with large angle errors.
As an alternative, we propose to use the decoupling metric from \eqref{eq:quat_decomp}, but proportional to the angle, as shown below.

\begin{align}
  \label{eq:propose1}
  e_R^{\textbf{G}}(R, \Rdes) = \rho_{tv} n_{tv} + \rho_{\text{yaw}} n_{\text{yaw}}
\end{align}

\input{comptable.tex}

Table~\ref{tab:comp} provides a tabulated summary of rotational error metric configurations discussed.

\input{experiments}

\section{Discussion}

We have shown that many quadrotor attitude controllers in the literature can be neatly characterized using their rotational error metrics' scaling and direction.

We have shown that rotational error metrics that decompose the attitude error in a thrust vector component and a yaw component provide superior control performance than those that use the full attitude error for (1) simultaneous large errors in position and yaw and (2) diagonal steps.

%
%
%
%
%
%

\bibliographystyle{IEEEtranN}
\bibliography{IEEEabrv,refs}

\end{document}

%% file: commands.tex
\DeclareMathOperator{\tr}{tr}

%% file: abstract.tex
\begin{abstract}
  We analyze and experimentally compare various rotational error metrics for use in quadrotor controllers.
  Traditional quadrotor attitude controllers have used Euler angles or the full rotation to compute an attitude error and scale that to compute a control response.
  Recently, several works have shown that prioritizing quadrotor tilt, or \textit{thrust vector} error, in the attitude controller leads to improved position control, especially in situations with large yaw error.
  We provide a catalog of proposed rotational metrics, place them into the same framework, and show that we can independently reason about and design the magnitude of the response and the direction of the response.
  Existing approaches mainly fall into two categories:
    (1) metrics that induce a response in the shortest direction to correct the full rotation error and
    (2) metrics that combine a response in the shortest direction to correct tilt error with the shortest direction to correct yaw error.
  We show experimental results to highlight the salient differences between the rotational error metrics.
  See \url{https://alspitz.github.io/roterrormetrics.html} for an interactive simulation visualizing the experiments performed.
\end{abstract}

%% file: comptable.tex
\begin{table}
  \begin{center}
  \caption{Comparison matrix of various rotational error metrics}
  \label{tab:comp}
  \begin{minipage}{\textwidth}
  \begin{tabular}{c|c|c}
    \textbf{Scaling / Direction} & \textbf{Full}: $n_e$ & \textbf{Decomposed}: $n_{tv} + n_{\text{yaw}}$ \\ \hline
    $\sin\rho$ &
      $e_R^{\textbf{A}}$: [\citenum{lee_geometric_2010,mellinger_minimum_2011,goodarzi_geometric_2013}] &
      $e_R^{\textbf{D}}$\footnote{For $\rho < \frac\pi2$.}: [\citenum{kooijman_trajectory_2019,gamagedara_geometric_2019}]
    \\ \\[-2.0mm] \hline \\[-2.8mm]
    $2\sin\frac\rho2$ &
      $e_R^{\textbf{B}}$: [\citenum{lee_exponential_2012,fresk_full_2013}] &
      $e_R^{\textbf{E}}$: [\citenum{brescianini_nonlinear_2013,faessler_automatic_2015,mueller_multicopter_2018,brescianini_tilt-prioritized_2020}]
    \\ \\[-2.0mm] \hline \\[-2.8mm]
    $\rho$ &
    $e_R^{\textbf{C}}$, $e_R^{\textbf{F}}$\footnote{For $\alpha_{\text{yaw}} > 0$.}: [\citenum{mueller_multicopter_2018}] &
      $e_R^{\textbf{G}}$: Proposed
  \end{tabular}
  \end{minipage}
  \end{center}
\end{table}

%% file: experiments.tex
\section{Experiments}

To evaluate the rotational error metrics for the purposes of quadrotor control, we use the following three scenarios.

\begin{enumerate}
  \item Direction change.
  \item Axis-aligned step with large initial yaw error.
  \item Diagonal step.
\end{enumerate}

For the last two experiments, we additionally compare against an Euler angle-based rotational error metric, defined using the Z-Y-X convention.
The yaw $\psi$, pitch $\theta$, and roll $\phi$, define a rotation matrix using the following map $F(\psi, \theta, \phi)$, with $c = \cos$ and $s = \sin$.

\begin{align}
  F = \begin{pmatrix} c\psi c\theta & c\psi s\theta s\phi - c\phi s\psi & s\psi s \phi + c\psi c\phi s\theta \\
  c\theta s\psi & c\psi c\phi + s\psi s\theta s\phi & c\phi s\psi s\theta - c\psi s\phi \\
-s\theta & c\theta s\phi & c\theta c\phi \end{pmatrix}
\end{align}

The rotational error metric is then defined using the inverse of $F$, making sure to compute the shortest angular distance for each angle.

\begin{align}
  e_R^{\textbf{ZYX}}(R, \Rdes) = F^{-1}(R) \ominus F^{-1}(\Rdes)
\end{align}

\subsection{Direction Change}

The quadrotor is tasked to reach a goal position at $(0, 3, 0)$ after starting at the origin with a velocity of $(0, -5, 0)$ and an initial roll angle of $80^\circ$ degrees.
This initial condition is designed to induce an initial roll error greater than $90^\circ$, so that differences in the rotational error metric responses at large angle errors are highlighted.

\begin{figure}
  \begin{center}
  \includegraphics[width=8cm]{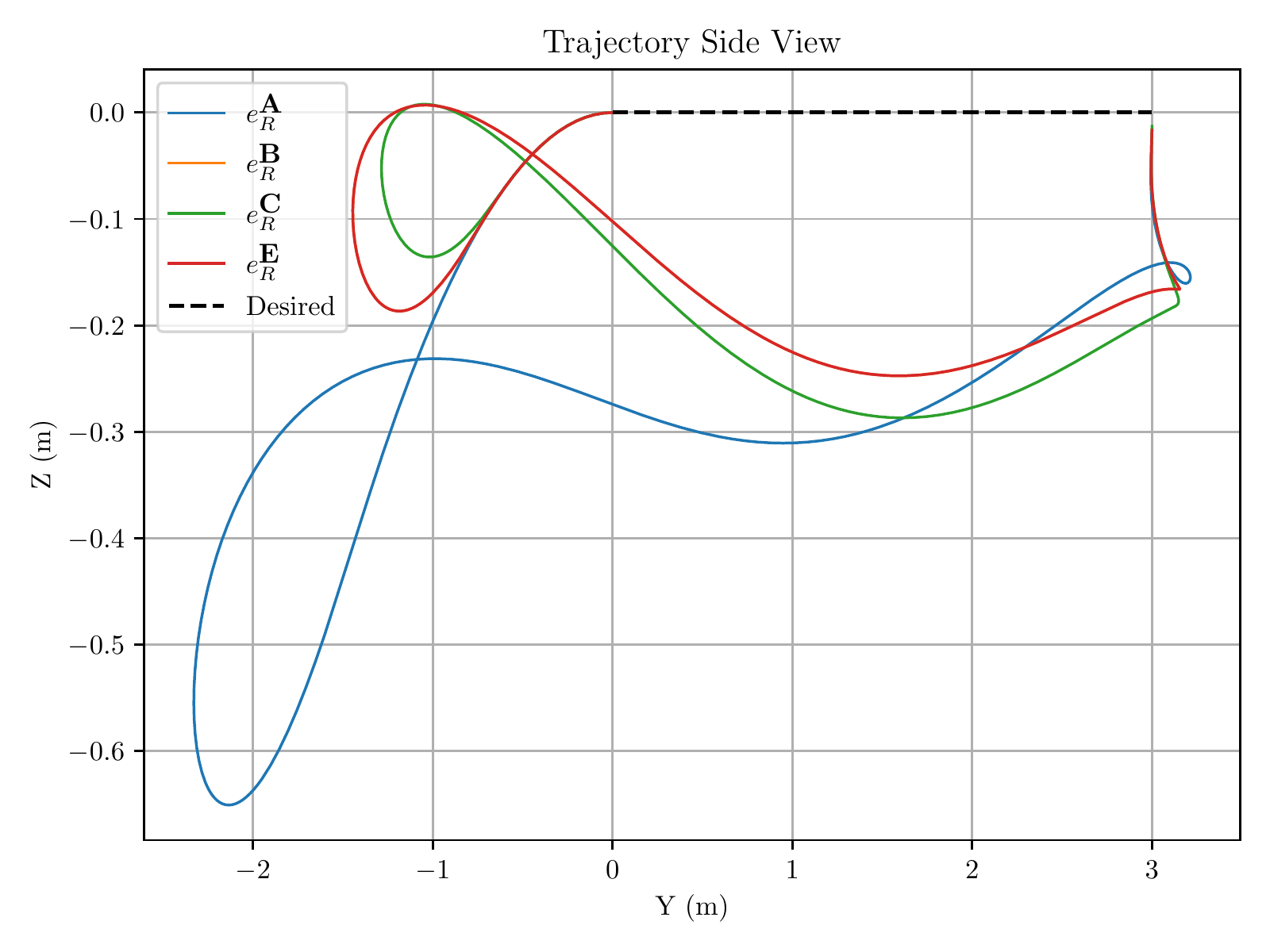}
  \includegraphics[width=8cm]{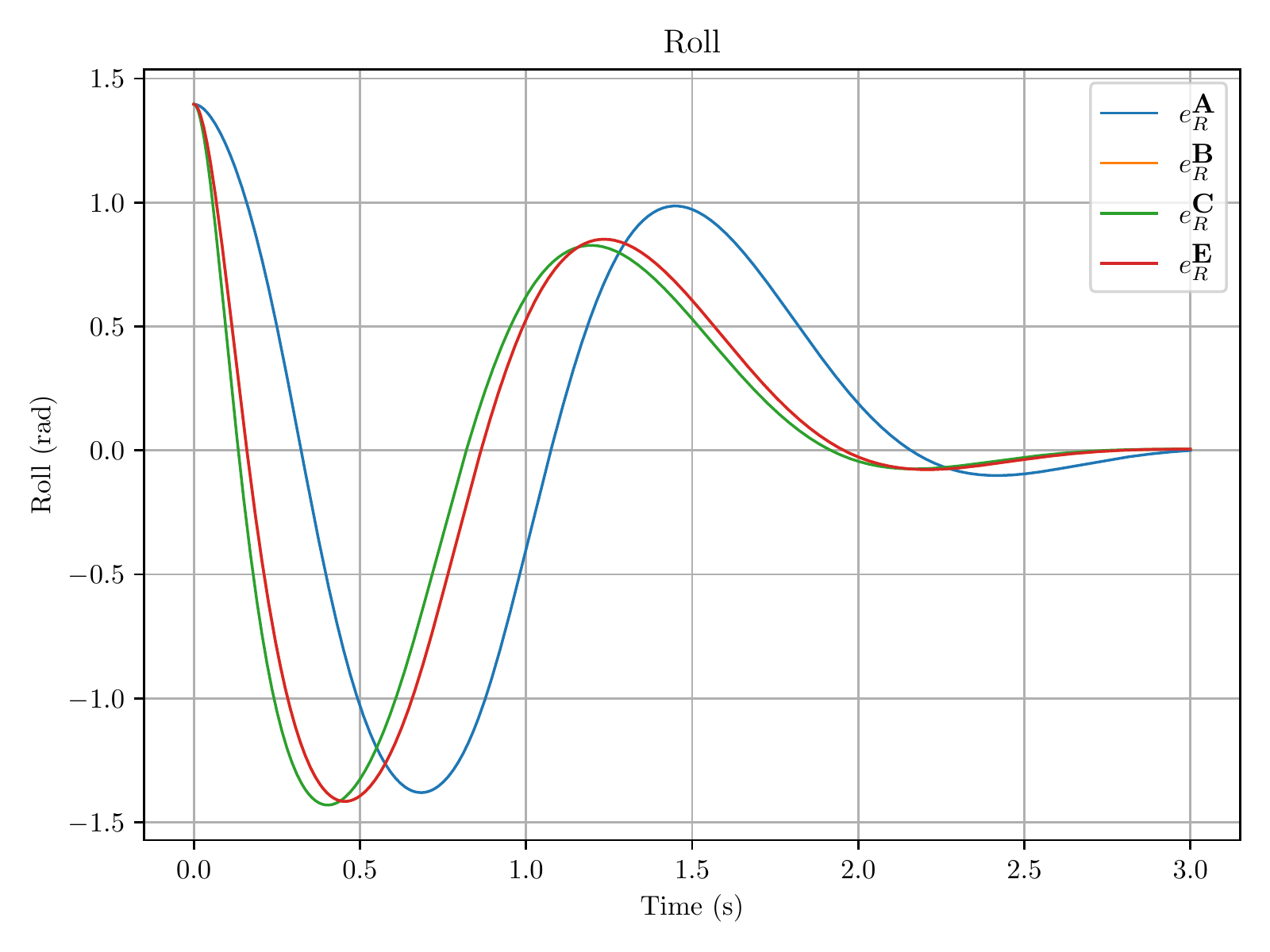}
  \caption{Side view of the trajectory followed (top) and roll (bottom) during the quick direction change test for various rotational error metrics. Metrics with a response proportional to the sine of the angle, \textbf{A}, converge much slower than those with a response proportional to sine of the half-angle, \textbf{B} and \textbf{E}, or the angle, \textbf{C}. Since the trajectory lies in the $YZ$ plane, \textbf{B} and \textbf{E} perform identically.}
  \label{fig:qc_side}
  \end{center}
\end{figure}

Figure~\ref{fig:qc_side} shows the resulting side view and roll trajectory.
The important thing to note here is that the rotational error metric $e_R^{\textbf{A}}$, which has a response proportional to the sine of the angle error, converges slower than the other metrics.
This example shows that metrics that induce a response proportional to the sine of the angle can suffer from slow convergence when there is large attitude error.

\subsection{Step with Yaw Error}

In this test, the quadrotor executes a position step and yaw step simultaneously.
The vehicle starts at the origin at a yaw of $90^\circ$ and moves to $(0, 3, 0)$ at a yaw of $0^\circ$.

\begin{figure}
  \begin{center}
  \includegraphics[width=8cm]{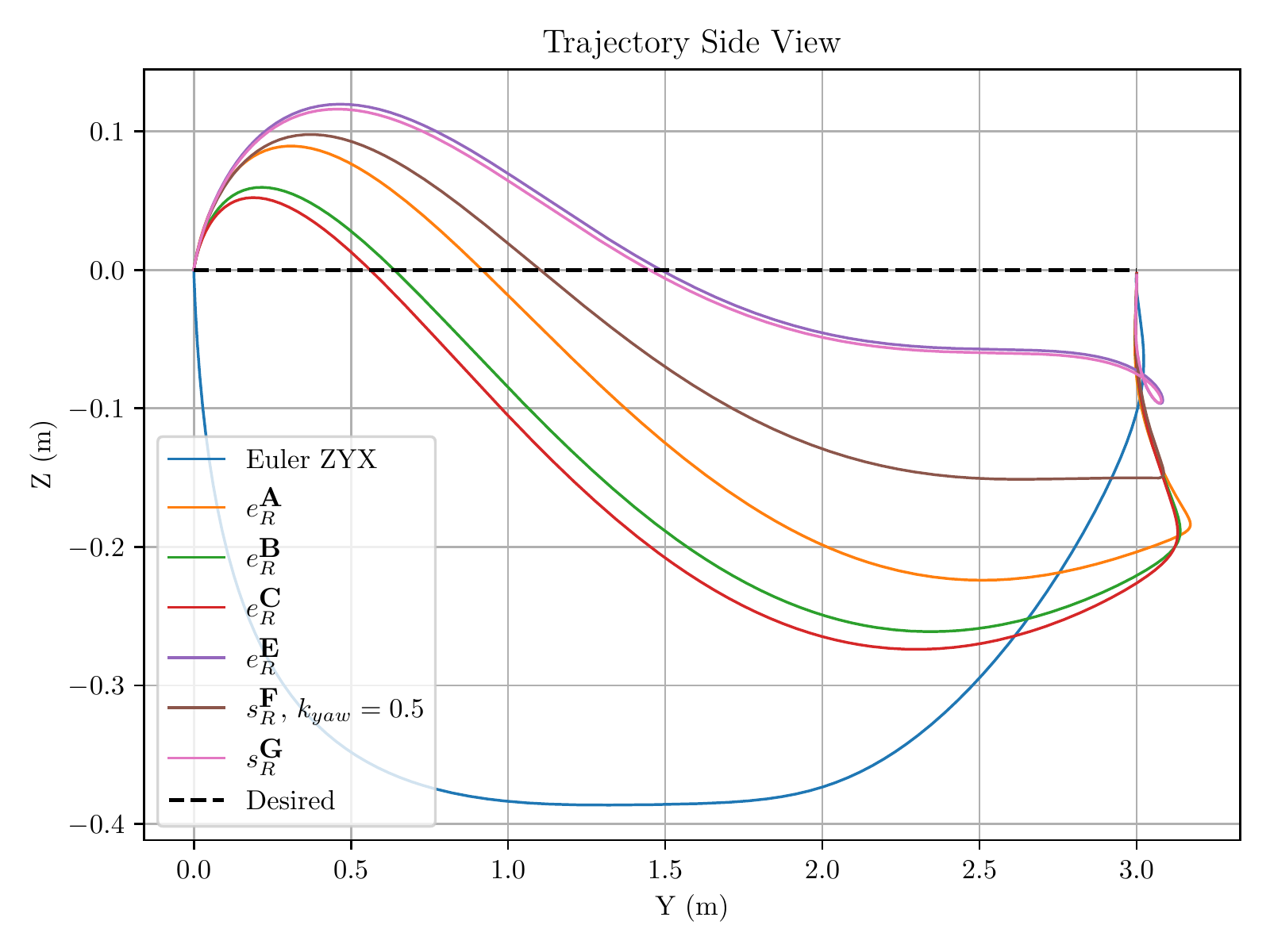}
  \includegraphics[width=8cm]{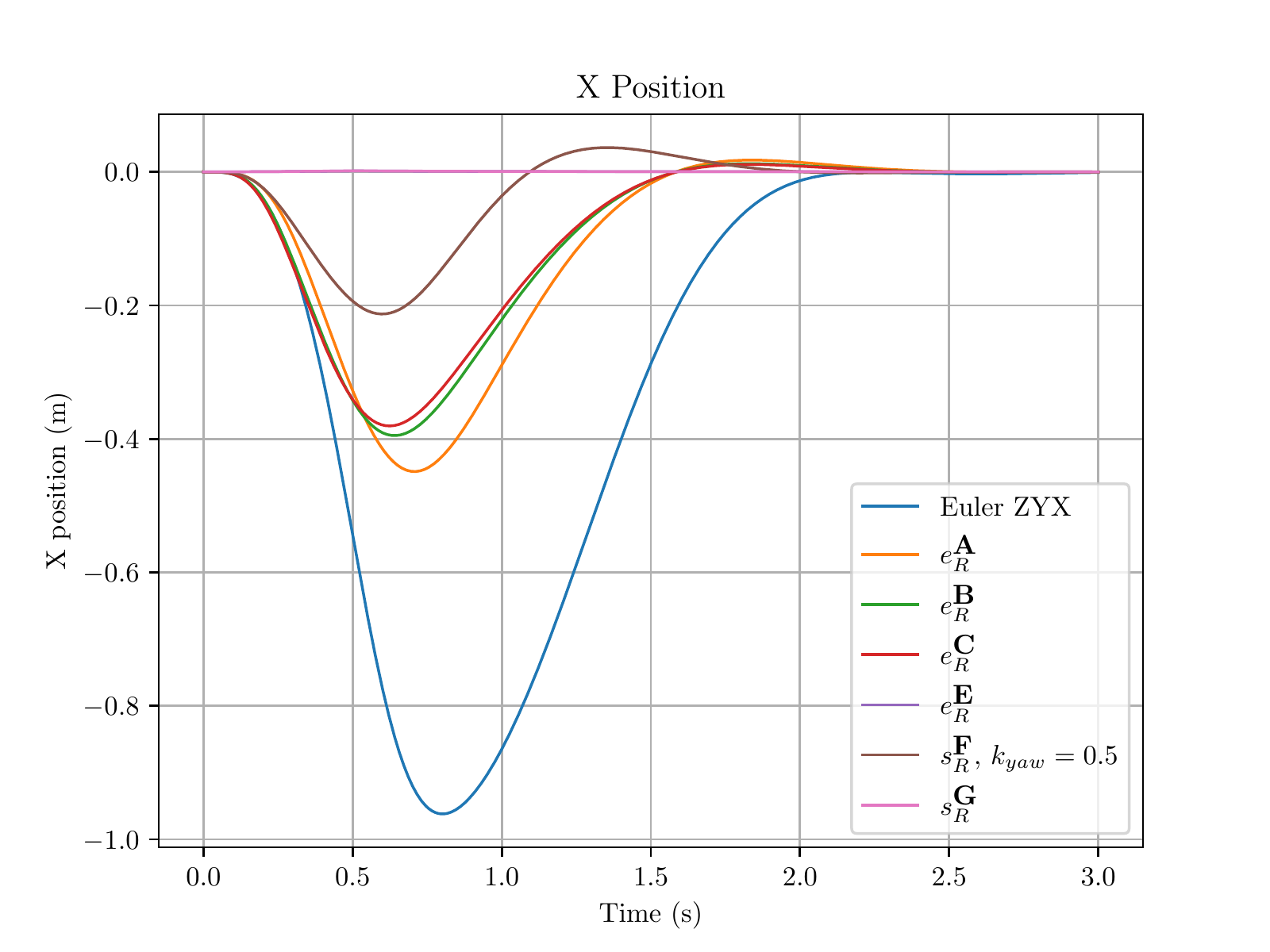}
  \caption{Side view of the trajectory followed (top) and $x$ position (bottom) during a simultaneous $3$ m position step and $90^\circ$ yaw step for various rotational error metrics. Only metrics that decompose the thrust vector from the yaw, \textbf{E} and \textbf{G}, maintain the desired $x$ position.}
  \label{fig:ys_side}
  \end{center}
\end{figure}

Figure~\ref{fig:ys_side} shows the side view and $x$ position during the experiment.
We see that metrics that use the full rotation error, result in trajectories that do not maintain the desired $x$ position.
Metrics that decompose the response, \textbf{E} and \textbf{G}, maintain the desired $x$ position.

\subsection{Diagonal Step}

In this test, we show that yaw error is not necessary for the rotational error metrics that do not decouple the thrust vector from yaw to exhibit suboptimal performance.
The quadrotor is given a diagonal step from the origin to $(3, 3, 0)$, with a desired yaw of $0^\circ$ for the duration of the test.

\begin{figure}
  \begin{center}
  \includegraphics[width=8cm]{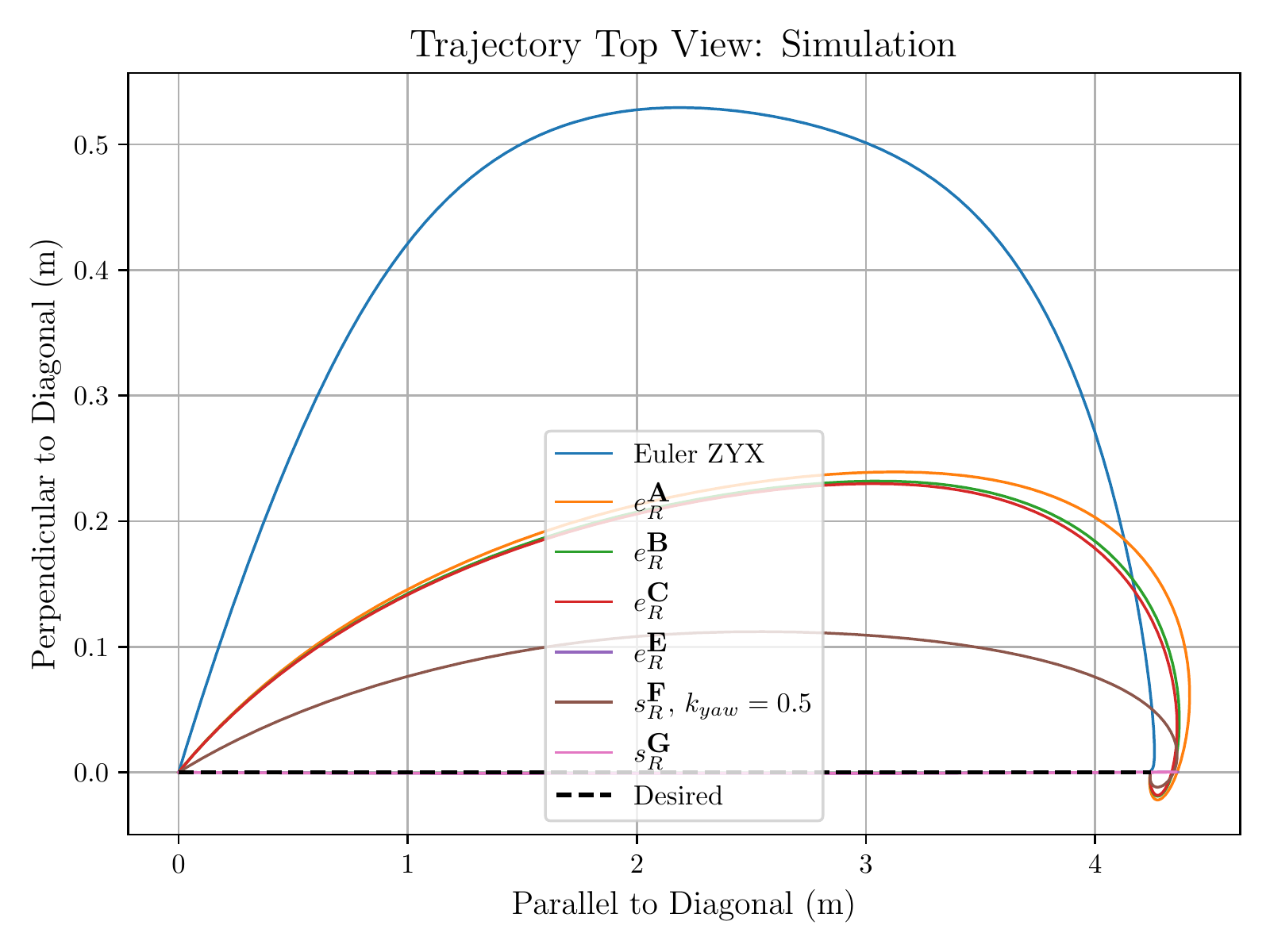}
  \caption{Top view of the trajectory followed during a diagonal step from the origin to $(3, 3, 0)$ for various rotational error metrics. Only metrics that decompose the thrust vector from the yaw, \textbf{E} and \textbf{G}, stay on the straight-line diagonal path.}
  \label{fig:ds_top}
  \end{center}
\end{figure}

Figure~\ref{fig:ds_top} shows the top view of the trajectories followed during the experiment for the various error metrics.
Metrics that use the full rotation error, \textbf{A}, \textbf{B}, \textbf{C}, and \textbf{F}, deviate from the diagonal path.
This is because the shortest path in $SO(3)$ from the identity to an orientation that results in the vehicle accelerating diagonally at the same yaw angle, takes the vehicle through orientations that accelerate in directions other than the diagonal direction. In other words, the thrust vector projected onto the horizontal plane does not point in the direction of the desired position.

Metrics that decompose the error into a thrust vector component and a yaw component, \textbf{E} and \textbf{G}, follow the straight-line diagonal path to the desired position.